# Scanflow: A multi-graph framework for Machine Learning workflow management, supervision, and debugging

Gusseppe Bravo-Rocca[a,∗], Peini Liu[a], Jordi Guitart[a,b], Ajay Dholakia[c], David Ellison[c], Jeffrey Falkanger[c] and Miroslav Hodak[c]

[a]*Emerging Technologies for Artificial Intelligence, Barcelona Supercomputing Center (BSC), Barcelona, Spain*
[b]*Computer Architecture Department, Universitat Politecnica de Catalunya (UPC), Barcelona, Spain*
[c]*Lenovo Data Center Group, Lenovo, Morrisville, NC, USA*

ABSTRACT

Machine Learning (ML) is more than just training models, the whole workflow must be considered. Once deployed, a ML model needs to be watched and constantly supervised and debugged to guarantee its validity and robustness in unexpected situations. Debugging in ML aims to identify (and address) the model weaknesses in not trivial contexts. Several techniques have been proposed to identify different types of model weaknesses, such as bias in classification, model decay, adversarial attacks, etc., yet there is not a generic framework that allows them to work in a collaborative, modular, portable, iterative way and, more importantly, flexible enough to allow both human- and machine-driven techniques. In this paper, we propose a novel containerized directed graph framework to support and accelerate end-to-end ML workflow management, supervision, and debugging. The framework allows defining and deploying ML workflows in containers, tracking their metadata, checking their behavior in production, and improving the models by using both learned and human-provided knowledge. We demonstrate these capabilities by integrating in the framework two hybrid systems to detect data drift distribution which identify the samples that are far from the latent space of the original distribution, ask for human intervention, and whether retrain the model or wrap it with a filter to remove the noise of corrupted data at inference time. We test these systems on MNIST-C, CIFAR-10-C, and FashionMNIST-C datasets, obtaining promising accuracy results with the help of human involvement.

## 1. Introduction

In recent years, despite the excellent results of Machine Learning (ML) approaches in different tasks such as machine translation (Bahdanau, Cho and Bengio, 2014; Luong, Pham and Manning, 2015), image classification (Krizhevsky, Sutskever and Hinton, 2017; Le, Zou, Yeung and Ng, 2011), recommendation systems (Covington, Adams and Sargin, 2016; He, Liao, Zhang, Nie, Hu and Chua, 2017), and speech recognition (Conneau, Baevski, Collobert, Mohamed and Auli, 2020), many concerns have arisen regarding the robustness of these techniques, such as security vulnerabilities (e.g., adversarial attacks, adversarial examples) (Takemura, Yanai and Fujiwara, 2020; Szegedy, Zaremba, Sutskever, Bruna, Erhan, Goodfellow and Fergus, 2014; Biggio, Corona, Maiorca, Nelson, Šrndić, Laskov, Giacinto and Roli, 2013; Kurakin, Goodfellow and Bengio, 2016), concept drift (e.g., class drift, covariate drift) (Gama, Žliobaitundefined, Bifet, Pechenizkiy and Bouchachia, 2014), lack of explainability and interpretability (e.g., black box networks) (Guidotti, Monreale, Ruggieri, Turini, Giannotti and Pedreschi, 2018; Goebel, Chander, Holzinger, Lecue, Akata, Stumpf, Kieseberg and Holzinger, 2018; Alvarez-Melis and Jaakkola, 2018), and maintainability or hidden technical debt (e.g., boundary erosion, unstable data dependencies, hidden feedback loops) (Sculley, Holt, Golovin, Davydov, Phillips, Ebner, Chaudhary, Young, Crespo and Dennison, 2015). These issues may occur at both design time (training) and run-time (inference).

From the previous examples, we can derive three ways in which a model can lose robustness. First, due to the data; for instance, if there is a change in the data distribution at run-time, the model drifts or behaves wrongly, which leads to accuracy degradation. Second, due to the model itself; for instance, a model with non-robust features that have highly predictive power but are brittle and incomprehensible to humans, does not generalize well in the presence of valid data perturbations and, as a result, is vulnerable to adversarial examples (Ilyas, Santurkar, Tsipras, Engstrom, Tran and Madry, 2019). Third, due to the workflow that encompasses the model; for instance, given a model that consumes data from a previous preprocessing step, if this input signal is unstable (can change its behavior over time), silent updates on the signal will affect the accuracy of the model.

Besides robustness issues at design and run-time, there are other convenient properties when building ML models to avoid the hidden technical debt, namely usability and scalability. Usability refers to how intuitive and efficient is to build workflows and interact with them, that is, how easy is that a human agent can be involved in the ML workflow to build, deploy, supervise, and improve models (Michael, Acklin and Scheuerman, 2020). Usability is also related

[∗]Corresponding author. Phone: +34651913568; Address: Carrer de Caballero, 10, 08014, Barcelona, Spain
gusseppe.bravo@bsc.es (G. Bravo-Rocca); peini.liu@bsc.es (P. Liu); jordi.guitart@bsc.es (J. Guitart); adholakia@lenovo.com (A. Dholakia); dellison@lenovo.com (D. Ellison); jfalkanger@lenovo.com (J. Falkanger); mhodak@lenovo.com (M. Hodak)
ORCID(s):

to portability and modularity. Portability allows dealing successfully with different libraries, operating systems, and any environment in general, when deploying workflows and, especially, when moving them between different platforms. Modularity allows adding new execution blocks to the workflow in a simple way. On the other side, scalability (Hestness, Narang, Ardalani, Diamos, Jun, Kianinejad, Patwary, Yang and Zhou, 2017) refers to the ability to handle increasing amounts of data and more expensive computations in an efficient way. Scalability in machine learning is necessary when dealing with huge data sets and complex models for both training and inference (Guo, Han, Yao, Wang, Xie and Yang, 2017). For instance, the ImageNet data set contains 1.2 million of images (Russakovsky, Deng, Su, Krause, Satheesh, Ma, Huang, Karpathy, Khosla, Bernstein, Berg and Fei-Fei, 2015), the real-time object detection model YOLO has 24 convolutional layers followed by 2 fully connected layers (Redmon, Divvala, Girshick and Farhadi, 2016), and the GPT-3 language model comprises 175 billion parameters (Brown, Mann, Ryder, Subbiah, Kaplan, Dhariwal, Neelakantan, Shyam, Sastry, Askell, Agarwal, Herbert-Voss, Krueger, Henighan, Child, Ramesh, Ziegler, Wu, Winter, Hesse, Chen, Sigler, Litwin, Gray, Chess, Clark, Berner, McCandlish, Radford, Sutskever and Amodei, 2020). Hence, the management of ML workflows must be able to exploit adequately resources in large clusters to handle their data and computations efficiently, as well as to allocate different types and amounts of resources to individual execution steps of the workflows according to their specific needs.

Consequently, there is a need for a framework that can help data scientists to define, deploy, track, supervise, and improve workflows in a flexible, usable, portable, modular, and scalable way so that they end up with more robust models. For that purpose, the framework has to combine machine-learned knowledge (to deal, for instance, with data distribution drift and data denoising) with human-provided knowledge (to deal, for instance, with bias, exceptions, fairness, and interpretability).

This paper contributes with a framework, so-called Scanflow, based on a graph structure where each node represents an executor (e.g., to handle tasks such as preprocessing, feature engineering, modeling, tuning), and each edge represents an interaction between nodes (e.g., input/output flux, constraints, addresses, and parameters in general). Each node runs its own image (which includes all the required software and dependencies) in a separate container to ease deployment, portability, reproducibility, and scalability.

The graph structure supports also three types of special nodes, which aim for workflow robustness by allowing to manage combined machine-learned and human-derived knowledge. In particular, each workflow can include its own Tracker, to log any metadata (settings, metrics, statistics, scores) for future analysis; Checker/s, to supervise the workflow behavior in production, identify abnormal observations, and allow human feedback; and Improver/s, to refine the model by using feedback from the Tracker and Checker nodes.

As a proof-of-concept, we have implemented a prototype of Scanflow including built-in versions of the Checker and the Improver nodes which feature a non-trivial anomaly detector and a denoising filter based on AutoEnconders that will detect and clean, respectively, out-of-distribution samples in the data and will improve a target estimator based on human intervention.

Scanflow, with its flexible graph structure and special built-in nodes (Trackers, Checkers, and Improvers), allows data scientists to set up reproducible dynamic workflows, track each execution, check for abnormal behaviors and, finally, refine the models based on both machine and human feedback.

The remainder of this paper is as follows. Section 2 states the formal definition of Scanflow, as well as its structure and collaborative elements, such as, graph design, deployment, and special nodes. Next, in Sections 3 and 4, we describe the Scanflow implementation and the results of our evaluation to demonstrate its usefulness when debugging ML models. Then, in Section 5 we present a comparison of different approaches regarding ML management, supervision, and debugging. Finally, the conclusions and new ideas for future work are presented in Section 6 .

## 2. Framework design

Latest ML algorithms have been focused mostly on data, learning, and computational power. However, terms such as knowledge and reasoning have been left out. As a consequence, many current models suffer from lack of generalization for unexpected or anomalous cases, i.e., samples that maintain the same semantics of the training data but are outside the ideal distribution of that set. Recent research work on the area has proposed the annexation of prior knowledge that can help to learn how to generalize better (Bengio, 2017). This previous work states that the manipulation of symbolic knowledge, for example, declarative knowledge (facts and rules) that contains statements with very high probability (i.e., ground truth), fits very well with the concept of conscious state. This is a step for the process of reasoning, which together with learning models, manages to help considerably in model generalization. A similar process occurs in the development of a baby's intelligence, where the mutual exclusivity of linguistic labels (symbolic system) works as a powerful tool to avoid overlapping concepts, allowing children to develop a rule-like generalization (Smith and Gasser, 2005). For instance, the word "ball" can be used for many objects that belong to this shape, similarly, the term "cup". In addition, the compositionality feature of language that allows children to understand a complex sentence by its components can aid in detecting and understanding the behavior of a function given a workflow, helping in the debugging process. We concur that any artificial intelligence or machine learning application has to be inspired by both symbolic human knowledge and the power of state-of-art learning algorithms.

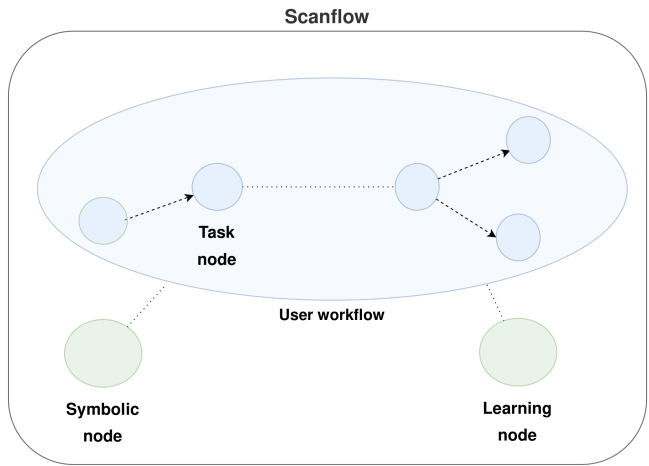

**Figure 1:** An abstract overview of Scanflow. A symbolic node represents a task node that involves human rules, knowledge base, or deterministic algorithms independently of experience. Whereas, a learning node depicts a pattern recognition behavior with the aim of fitting or approximating a function, like deep learning approaches that depend on experience. The goal of wrapping a user workflow with new nodes is to increase its robustness.

Consequently, we propose Scanflow, a multi-graph framework that combines both machine (learning nodes) and human knowledge (symbolic nodes) to tackle the problem of model robustness. On one side, using machine learning will help to detect any abnormal behavior in a workflow with ease. On the other side, with the help of human knowledge we can tackle the bias in algorithms by supervising the type of data that is fed to the model or creating rule-based procedures to guarantee an expected behavior. Scanflow involves humans in creating models that other humans can trust. In this section, we outline the implementation of the Scanflow graph structure that allows both approaches to work side by side. Figure 1 depicts this idea.

## 2.1. Definition

Scanflow is a directed labeled multigraph represented by a 7-tuple
$G = (\Sigma_V, \Sigma_E, V, E, f, \ell_V, \ell_E)$, where

$\Sigma_V$ is the alphabet of vertex labels. $V$ is the set of nodes or vertices. Each node is considered a task node which performs some action.

$\Sigma_E$ is the alphabet of edge labels. $E$ is the set of ordered edges. Each edge represents a dependency between nodes, that is, which nodes should be executed before or after others.

$f : E \to V \times V = \{(u,v)|u,v \in V\}$ is the function that maps each edge to a pair of vertices or nodes. $f(e_1) = f(e_2)$ in case of multiple edges.

$\ell_V : V \to \Sigma_V$, maps a label for each node. For instance, a node can be a preprocessing node, cleaning node or modeling node belonging to a certain workflow.

$\ell_E : E \to \Sigma_E$, maps a label for each edge. For instance, an edge (e.g., gathering → preprocessing) can be labeled as belonging to the training phase or belonging to the inference process.

There are three special nodes: $\{T, C, I\} \in V$, *Tracker* (T), *Checker* (C), and *Improver* (I), which are instances of symbolic and learning nodes that will interact with each other to enhance either a node or set of nodes. A *Tracker* node is a symbolic node since it depicts a declarative knowledge (facts), for instance, logging any execution metadata. A *Checker* node can behave both as a learning or as a symbolic node since it takes both pattern recognition and rule-based ideas. Finally, an *Improver* node shares also both behaviors, symbolic (e.g., rules to decide when to update the model) and learning (e.g., retrain the model based on feedback from *Tracker* and *Checker*).

Finally, let be $\mathcal{R}(G) = \sum_{k=0}^{n} i_G^{(k)}$ a return function that measures all the improvements $i$, jobs done by the *Improver* nodes, applied to a graph $G$. Then, the goal is to get a graph $G_\theta^*(w)$ for a given workflow $w(x)$ (graph that depends on an input $x$) and metric $\theta$ of quality (e.g., accuracy) that maximizes $\mathcal{R}$, where

$$G_\theta^*(w) = \arg\max_\theta \mathcal{R}(G_\theta(w)) \qquad (1)$$

$$w(x) = [n_{(x)}^{(0)}, n_{(x)}^{(1)}, \ldots, n_{(x)}^{(k-1)}] = \begin{cases} n^{(0)} & \text{1th task node} \\ n^{(1)} & \text{2th task node} \\ \vdots \\ n^{(k-1)} & \text{kth task node} \end{cases} \qquad (2)$$

As a first attempt to implement Scanflow, we propose a blueprint comprising five modules: *Setup* and *Deployment* to define and execute a workflow, respectively, a *Tracking* module to log or save metadata in any execution, a *Checking* module to tag any observation, and finally, an *Improving* module to apply all the collected feedback coming from whether *Tracking* or *Checking* modules to be used to enhance the target model. Figure 2 depicts a running state of this template.

## 2.2. Modules
### 2.2.1. Setup

This module is responsible for designing the workflow or graph in which each node will be assigned a specific task, e.g., preprocessing, cleaning, modeling, etc. Additionally, special nodes (Tracker, Checker, and Improver) can be also defined, so that, together with the existing user nodes, a more robust workflow can be obtained.

Apart from the definition of the graph, it is also possible to define the order of execution of each node, that is, which nodes should be executed before or after others. In particular, Scanflow allows sequential and parallel execution of nodes. This provides a greater flexibility when working with ML workflows in the training and inference phases, since the user has total control of each node and of the information transmitted in each path (nodes in the same neighborhood). This is useful when one wants to create multiple versions (reuse) of the same node with different parameters and execute them in a parallel way. This could reduce training times, help in the modularization, or simplify the experimentation with specific cases that in a standard execution would be very expensive and cumbersome. Figure 3 shows an example of this idea, where two workflows are connected to one another. The first workflow, which runs sequentially, can include a data extraction node, which precedes a cleaning node, which then precedes a feature engineering node. The second workflow can comprise a modeling node, which executes models written in Pytorch or Tensorflow or Sklearn in a parallel way and with different arguments. Additionally, note that each workflow can have a Checker and/or a Tracker node.

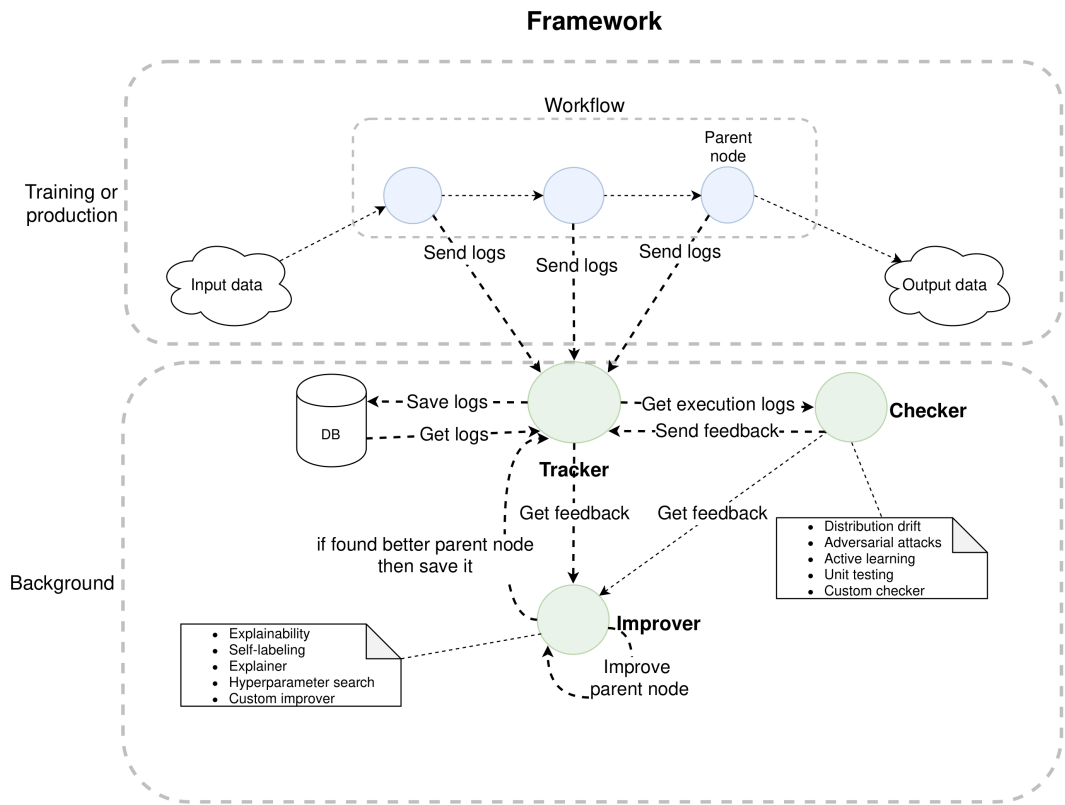

**Figure 2:** Scanflow execution cycle. Figure shows the interaction between Tracker, Checker, and Improver nodes to enhance the model quality once the workflow is already established.

### 2.2.2. Deployment

This module is in charge of deploying each task node within a container (which has more capabilities than a simple Python virtual environment, e.g., allocating different Linux kernels, compilers, etc.), with the purpose of encapsulating it in an isolated environment including its own software dependencies, its view of the operating system, and its computational resources, which provides advantages on modularization, portability, and reproducibility. Conceptually, our design allows the containers to be implemented in any containerization technology, although our prototype is currently built on top of Docker (as described later). Once the containers are in place, they are grouped into a virtual network (e.g., with private IP addresses) to enable isolated communication between them. This can be seen in Figure 4. Additionally, this module executes the workflow in the environment where it has been deployed (in local, in a server, or in a cluster). Once the workflow is running, along with the special nodes (Tracker/s, Checker/s, Improver/s), Scanflow allows to proceed with the testing or debugging of the workflow to improve its robustness.

### 2.2.3. Special nodes

The two previous modules were responsible for designing and deploying the workflow through a graph of containers to have a flexible structure which solves the problems of usability (simplifying deployment, portability, and reproducibility) and scalability (using containers allows to finetune the resources to work with large data and more complex models). In spite of these benefits, the real advantage for robustness lays down in the definition of the special nodes, namely the Tracker, the Checker, and the Improver, which are described below.

### 2.2.4. Tracking

The Tracking module is in charge of saving any kind of metadata coming from the workflow (e.g., parameters, metrics, statistics, split datasets, etc.), as well as metadata coming from other special nodes, such as the Checker (e.g., anomalous points) and the Improver (e.g., new model, new score). The purpose of the Tracker is to offer this information to the other special nodes so that they can improve the final model or even the workflow as a whole. The more information obtained from the execution of the workflow, the better for the performance of the other special nodes. Figure 5 shows a high-level overview of this module. Furthermore, the way of how this module works is shown in Algorithm 1, which saves metadata to the Tracker, and Algorithm 2, which queries the Tracker for specific metadata.

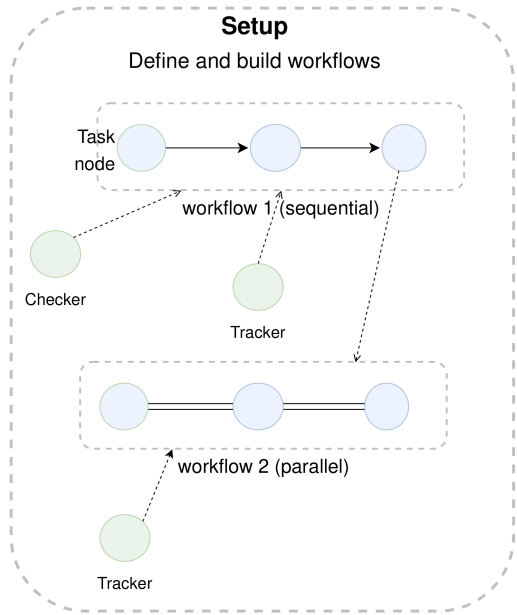

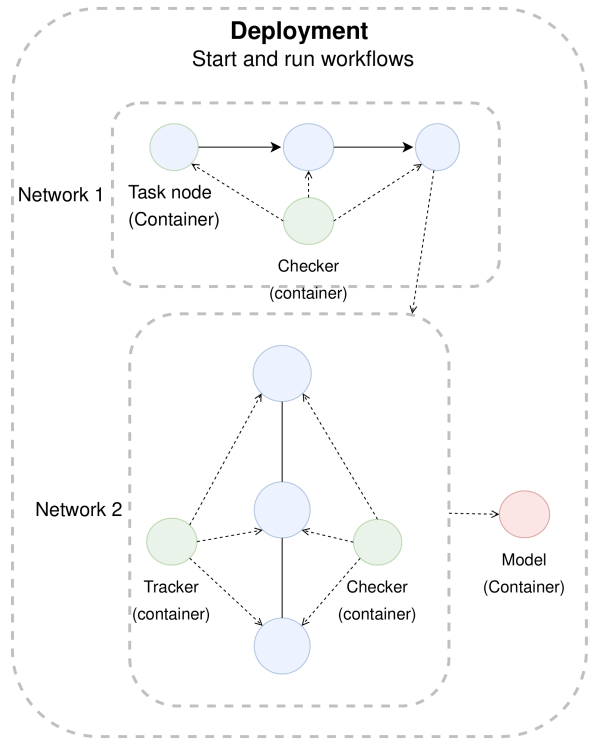

**Figure 3:** *Setup* module. The blue nodes belong to the current user workflow, whereas the green ones represent the special nodes that will help to increase the quality and reliability. Besides, this figure shows one graph comprising two subgraphs, one with a sequential order (top) and one with parallel order (bottom).

**Figure 4:** *Deployment* module. Once the workflow is defined (in both structure and order), this module uses containers to deploy and run this graph, i.e., execute the task and special nodes.

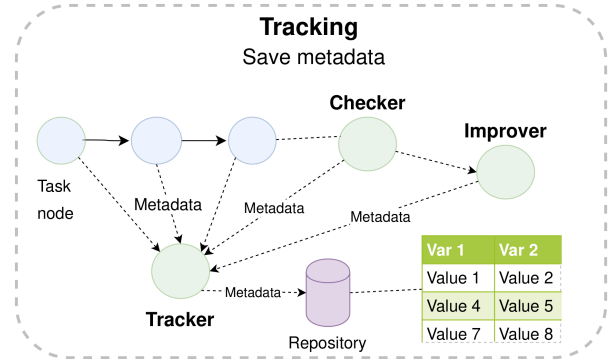

**Figure 5:** *Tracking* module. The user's workflow (blue nodes), the Checker (green node), and the Improver (green node) send metadata to the Tracker node to be used later to refine the model. This metadata will be saved in a shared or private repository.

---

**Algorithm 1:** High-level *tracking* algorithm for saving metadata

**Input**: $id$: execution identifier, $meta$: metadata;
// Process metadata
pmeta ← process_metadata(meta);
// Save processed metadata to repository
flag ← save_metadata(pmeta, id);
return flag;
**Output**: $flag$: success or failure;

---

### 2.2.5. Checking

To obtain a robust workflow, its behavior over time must be analyzed in order to detect anomalous observations and apply corrective measures. The Checking module is in charge of identifying different types of behaviors through predefined instructions. Figure 6 depicts this idea. To this end, Checker nodes can be built to implement symbolic or learning behaviors. A symbolic type node can hold rules or some knowledge in order to help the model decisions. For example, some rules can constrain the minimum and maximum temperatures a sensor can measure, so that, if the measurement exceeds 1000 Celsius degrees or drops below -273 Celsius degrees, it is considered to be caused by a sensor malfunction. Another example of a symbolic node is human labeling, where a human agent labels observations to be used later for refining the model with an extended training set. This can contribute to a great extent to the generalization of the model. On the other hand, a learning type node is

**Algorithm 2:** High-level *tracking* algorithm for getting metadata

**Input**: *id*: execution identifier, *query*: request;
```
// Get tracker address given an experiment id
```
tracker_uri ← get_tracker_uri(id);
```
// Get tracker node instance given a tracker
   address
```
tracker ← get_tracking_uri(tracker_uri);
```
// Gather data from tracker given a query
```
data ← tracker.gather_log(query);
return data;
**Output**: *data*: requested information;

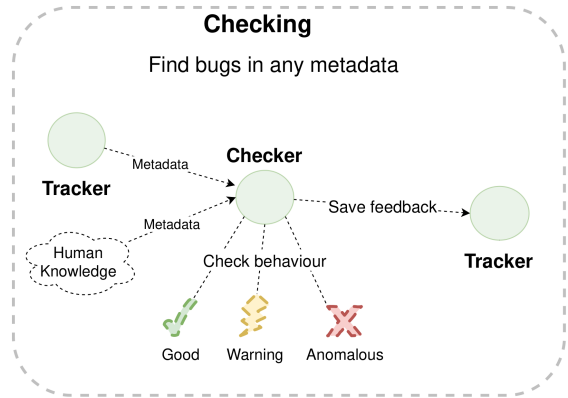

**Figure 6:** *Checking* module. It is in charge of analyzing some data and deciding if it behaves as expected or not. It can add a tag indicating whether the data is anomalous or not, or some continuous severity level.

in charge of predicting anomalous behavior on some data by using patterns learnt from historical data. For example, a classification model of dogs and cats can fail if new images fed into the model, in the inference stage, deviate from the training distribution (e.g., cats with dog breeds, big cats, small dogs). A general formulation of the Checking module is shown in Algorithm 3.

**Algorithm 3:** High-level *checking* algorithm

**Input**: $X$: metadata;
feedback ← *None*;
```
// Prepare the raw metadata
```
X_prep ← preprocessing(X);
```
// Call a symbolic or learning function to check
   behaviour
```
tags ← tagger(X_prep);
```
// Save tags for later analysis
```
feedback ← tags;
return feedback;
**Output**: *feedback*: information for improvement;

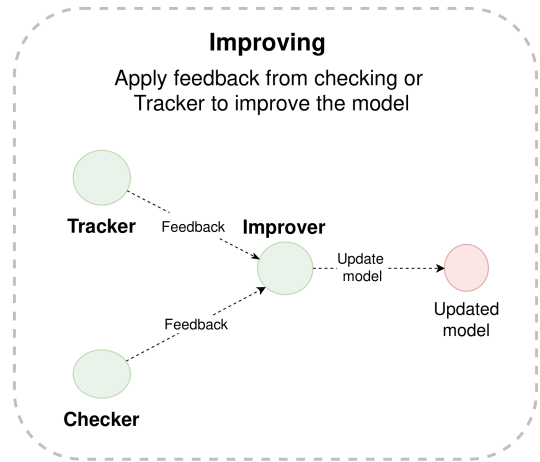

**Figure 7:** *Improving* module. The feedback coming from different sources is used to enhance the current model. The updated model is shown in red color.

### 2.2.6. Improving

The Improving module is in charge of carrying out the corrective measures to the model (in a general way, it could improve any node of the workflow, not only the model). As shown in Figure 7, the information obtained from the Tracker and/or Checker will be used to address anomalous behaviors not seen in the training stage with the aim of optimizing some given metric, such as, accuracy, execution time, interpretability, trust, etc. To this end, the Improver node takes the feedback from the other nodes and refines the current model, obtaining a new version of it. After this step, it measures how much the improvement is with respect to the old model, and if it is higher than a given threshold, then the new model is saved for a future update triggered by a previous authorization (human intervention) or in an unsupervised way. Algorithm 4 shows the blueprint of this module.

**Algorithm 4:** High-level *improving* algorithm

**Input**: $m$: model;
**Input**: *feedback*: any kind of metadata;
**Input**: $q$: current quality metric;
```
// Modify or create a wrapper for the current
   model
```
$m'$ ← modify($m$, feedback);
```
// Get score from modified model
```
$q'$ ← evaluate($m'$);
```
// If new model is better, save it, else, do
   nothing
```
**if** $q' > q$ **then**
| return $m'$, $q'$;
**end**
return None, None;
**Output**: $m'$: new model;
**Output**: $q'$: new quality value;

Gusseppe Bravo-Rocca et al.: *Preprint submitted to Elsevier*     Page 6 of 22

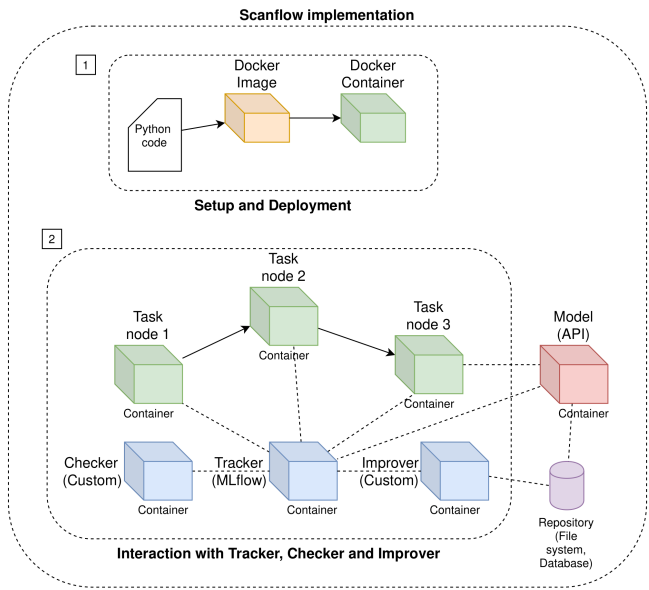

Figure 8: Scanflow's structure. This is a proof-of-concept implementation using Docker, MLflow, and Python. At the top, the core structure is shown; each task node is encapsulated into a Docker image, which is then instantiated as a container. At the bottom, the interaction of user task nodes and special nodes is shown.

## 3. Implementation

### 3.1. Graph structure

Scanflow is implemented under the principles of graph theory and containerization, which provide the needed support to enable ML model debugging. As for this first stage, Scanflow is written in Python on top of Docker (in the future, we plan to extend it to other technologies), which allow easy integration with a variety of libraries and task isolation, respectively. Additionally, Scanflow structure facilitates fast prototyping, easy deployment, portability, reproducibility, and scalability.

The encapsulation of any node (e.g., preprocessing, modeling, cleaning) will create a Docker image (which includes all the required software and dependencies) at the Setup stage. This image will then be instantiated into a Docker container. The whole process is shown in Figure 8. As a result, a workflow is represented by a set of containers and running this workflow (Deployment stage) means executing each one of the containers according to the order defined in the Setup. It should be noted that the ML task encapsulated in each image can be written in any language and use any library, since its ecosystem is isolated from other nodes. Furthermore, Scanflow provides a registry of all the images for later reuse.

### 3.2. Graph communication

The communication between nodes is a key point in the interaction within the workflow. This gives the characteristic of 'multi' to a graph, i.e., the possibility of using one or more edges between two nodes. As shown in Figure 9, the communication between two nodes in Scanflow is done in two ways. The first one is through a Docker shared volume. Our proof-of-concept implementation sets a space in the file system to persist the data, which can be shared by any container running in the same host. This type of communication is faster for files that need to be used many times. Although not part of the current implementation, Docker supports also volume drivers that allow storing volumes on remote hosts or Cloud providers. The second communication option is through a Tracker, which receives queries and delivers the results from its database. This type of communication serves for metadata, such as logs, metrics, scores, parameters, etc., and allows communicating containers in both local and remote hosts.

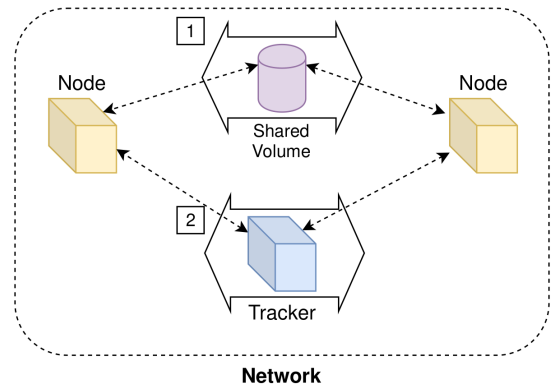

Figure 9: Scanflow's communication between nodes. The first link is using a Docker shared volume and the second is using a Tracker. These nodes can be either a task node or a special node.

### 3.3. Special nodes

The special nodes implement tracking, checking, or improving functionality. Each of them features machine-learned or human-derived knowledge to improve workflow robustness. The Tracker node, for example, is implemented using MLflow[1], an open source library that helps in event tracking and metadata management, but also in data acquisition with predefined methods. The Checker and Improver nodes run on-demand (or in the background), observing the data that gets into the model, detecting anomalous events, and trying to improve the model using information from those events. The implementation of the Checker and the Improver nodes depends on the specific functionality they are supporting (for instance, dealing with model bias, fairness, and interpretability). Next section describes the built-in implementation of those nodes included in Scanflow.

### 3.4. Built-in special nodes

Scanflow comes with novel built-in Checker and Improver nodes which aim to detect data drift distribution and refine the model or the out-of-distribution data accordingly by using human feedback. This is motivated by the need of being aware of any deviation of the data that leads to a malfunction of the model. For example, if a model is asked to classify an image, but this image contains some noise, in many cases the model fails and predicts a totally wrong class, which means that the model is not robust enough to generalize to other scenarios.

---

[1]https://mlflow.org/

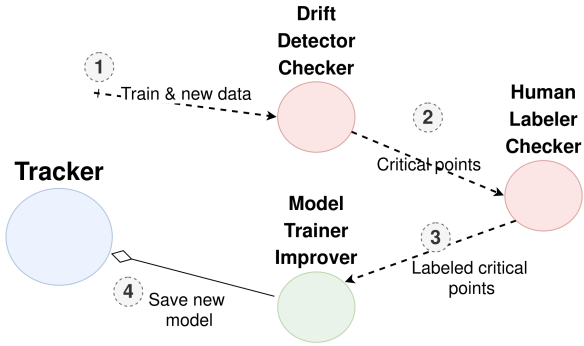

**Figure 10:** First ML debugging approach. Joint work between Drift Detector Checker, Human Labeler Checker, and Model Trainer Improver.

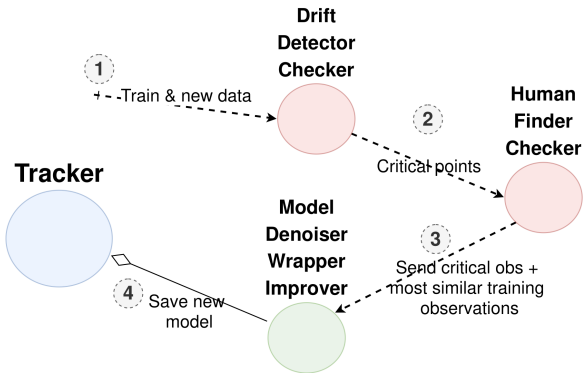

**Figure 11:** Second ML debugging approach. Joint work between Drift Detector Checker, Human Labeler Finder, and Model Denoiser Wrapper Improver.

We explored two approaches to deal with this issue. The first one, as shown in Figure 10, comprises three components, namely a Drift Detector Checker, which analyzes the out-of-distribution observations that occur in the inference stage by calculating the distance between the new data and the original data, a Human Labeler Checker, which asks a human agent to label the further away samples that indicate non-expected behaviors, and a Model Trainer Improver, which retrains the model with the extended training data to improve a target estimator. The second approach, which is shown in Figure 11, aims to remove the noise from the corrupted input data. For this purpose, it first uses the same Drift Detector Checker to detect the corrupted observations, then a Human Finder Checker looks for clean observations in the training data similar to the corrupted ones, and finally, a Model Denoiser Wrapper Improver connects a denoiser to the current model in order to filter out the noisy input before predicting. The detailed implementation of these two approaches is described in the next sections.

### 3.4.1. Background

According to Webb, Hyde, Cao, Nguyen and Petitjean (2016), there are many types of concept drift, such as class drift, covariate drift, novel class drift, etc. For this work, we consider the covariate drift, where the features ($X$) or covariates change over time. A more formal definition is as follows. Let be $P(X)$ the prior probability distribution over the covariates, $P(Y)$ the prior probability distribution over the class labels, $P(Y \mid X)$ the likelihood distribution over class labels given $X$, and $P(X \mid Y)$ the posterior probability distribution over covariates given class labels, then a covariate drift is defined as shown in Equation 3. It means that for two times $t$ and $u$, the prior probability distribution changes for each case, as long as, posterior class distribution remains the same.

$$P_t(X) \neq P_u(X) \land P_t(Y \mid X) = P_u(Y \mid X) \qquad (3)$$

Equation 3 tells us that by maintaining the semantics of the classification, that is, predicting the same $Y$, the values ($X$) that feed into the model have changed over time, forcing the model to readjust in order to deal with this distribution variation. One way to detect this variation is through the idea of coding, that is, building an architecture capable of learning internal representations through a coding process. Rumelhart, Hinton and Williams (1986) used a neural network in which the input and output layer have the same $N$ dimension and the hidden layer has a $\log_2 N$ dimension. Their idea is to code an $N$ bit pattern into a $\log_2 N$ bit pattern, and then decode that representation into the output $N$ bit pattern. Lecun (1987) and Bourlard and Kamp (1988) came up with the idea of *autoencoder*, an architecture made up of two components: an encoder and a decoder. The encoder is a function $f$, as shown in Equation 4, that allows to code the input data into a specific representation called latent space $h$. Conversely, a decoder maps $g$, as shown in Equation 5, from a latent space $h$ to input space, which produces a reconstruction $r$ that mimics the input space $x$ (Bengio, Courville and Vincent, 2013). The following equations show the backbone idea of a general autoencoder:

$$h = f_\theta(x) \qquad (4)$$

$$r = g_\theta(h) \qquad (5)$$

$$\mathcal{J}_{AE}(\theta) = \sum_t L\left(x^{(t)}, g_\theta\left(f_\theta\left(x^{(t)}\right)\right)\right) \qquad (6)$$

Here $\mathcal{J}_{AE}$ (Equation 6) represents the loss overall function that penalizes $r$ for being dissimilar from $x$. Figure 12 shows a general structure of an autoencoder. Similarly, a denoising autoencoder (DAE) (Vincent, Larochelle, Bengio and Manzagol, 2008) goes one step further. It changes the loss function adding a corrupted version ($\hat{x}$) of the input data ($x$) obtained through a corruption process $C(\hat{x}|x)$. This modified loss function, as shown in Equation 7, forces the autoencoder to remove the noise and obtain useful latent features (not only mapping or learning the identity function). Figure 13 depicts this idea.

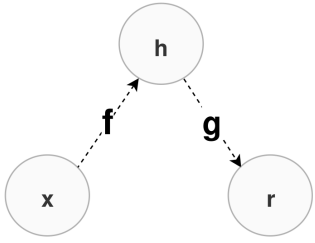

**Figure 12:** Naive autoencoder

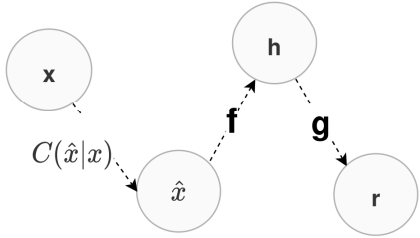

**Figure 13:** Denoising autoencoder

$$\mathcal{J}_{DAE}(\theta) = \sum_t L\left(x^{(t)}, g_\theta\left(f_\theta\left(\hat{x}^{(t)}\right)\right)\right) \quad (7)$$

### 3.4.2. Drift Detector Checker

The Drift Detector Checker is capable of capturing deviated points using only the data considered the "normal distribution", that is, the training data. In an unsupervised manner, this detector is trained on this set to learn a function to reconstruct the new data into the original one. Using this reconstruction error, the detector is able to calculate the anomalous or critical points.

As shown in Figure 14, our built-in Drift Detector Checker consists of two parts, namely a Convolutional Deep Autoencoder and a Critical Point Selector, and, as shown in Algorithm 5, its logic follows the high-level blueprint presented before for Checker nodes.

The Convolutional Deep Autoencoder, which contains 7 convolutional layers, 3 max-pooling layers, and 3 up-sampling layers, calculates how far the new points are from the original distribution. This is shown in Algorithm 6.

Then, by using this result, the Auto-threshold component of the Critical Point Selector calculates two thresholds, one upper and one lower, in order to identify the points that are out of these boundaries. Given these limits, the Kernel Density Estimator component of the Critical Point Selector estimates the distribution density of these critical points from the previous step, which is then used to choose the

---

**Algorithm 5:** Drift Detector Checker

**Input**: $X\_train$: training data, original distribution;
**Input**: $X\_test$: testing data, data that contains corruptions;
feedback ← $None$;
// Learn the original distribution
autoencoder ← conv_autoencoder(X_train);
// Prepare the testing data, like normalization, scaling, etc.
X_test_prep ← preprocessing(X_test);
// Find the observations that deviate the original distribution
X_critical ← crit_point_selector(autoencoder, X_train, X_test_prep);
// Save the most critical observations
feedback ← X_critical;
return feedback;
**Output**: $feedback$: critical observations, subset of X_test;

---

**Algorithm 6:** Convolutional Deep AutoEncoder

**Input**: $X$: training data;
**Input**: $epochs$, $batches$, $lr$: Number of epochs, number of batches, learning rate;
**Input**: $W$, $b$, $b_h$: Parameters ($\theta$) : Weights, bias, latent bias;
**for** $epoch\ in\ epochs$ **do**
    **for** $batch\ in\ batches$ **do**
        $h \leftarrow \text{relu}(\boldsymbol{X} * \boldsymbol{W} + \boldsymbol{b})$
        $r \leftarrow \text{sigmoid}\left(\boldsymbol{h} * \boldsymbol{W}^T + \boldsymbol{b_h}\right)$
        $L(\boldsymbol{x}, \boldsymbol{r}) \leftarrow$
        $-\sum_{i=1}^d \left[\boldsymbol{x}_i \log \boldsymbol{r}_i + (1 - \boldsymbol{x}_i) \log (1 - \boldsymbol{r}_i)\right]$
        $g \leftarrow \text{gradient}(\theta)$
        **for** $\theta_i, g_i\ in\ (\theta, g)$ **do**
            $\theta_i \leftarrow \theta_i - l * g_i$
        **end**
    **end**
**end**
autoencoder ← join(h, r)
return autoencoder
**Output**: $autoencoder$: convolutional deep autoencoder;

---

required number of the most likely critical points. The logic of the Critical Point Selector, including the Auto-threshold and the Kernel Density Estimator, is shown in Algorithm 7.

### 3.4.3. Human Labeler Checker

This Checker node is a type of symbolic node, that is to say, it can be anything that does not depend on the historical data: a rule, a fact, or a human intervention. In this case, it is a human that works as a participatory agent within this multi-node environment. His or her job is to take the observations

Gusseppe Bravo-Rocca et al.: *Preprint submitted to Elsevier*  Page 9 of 22

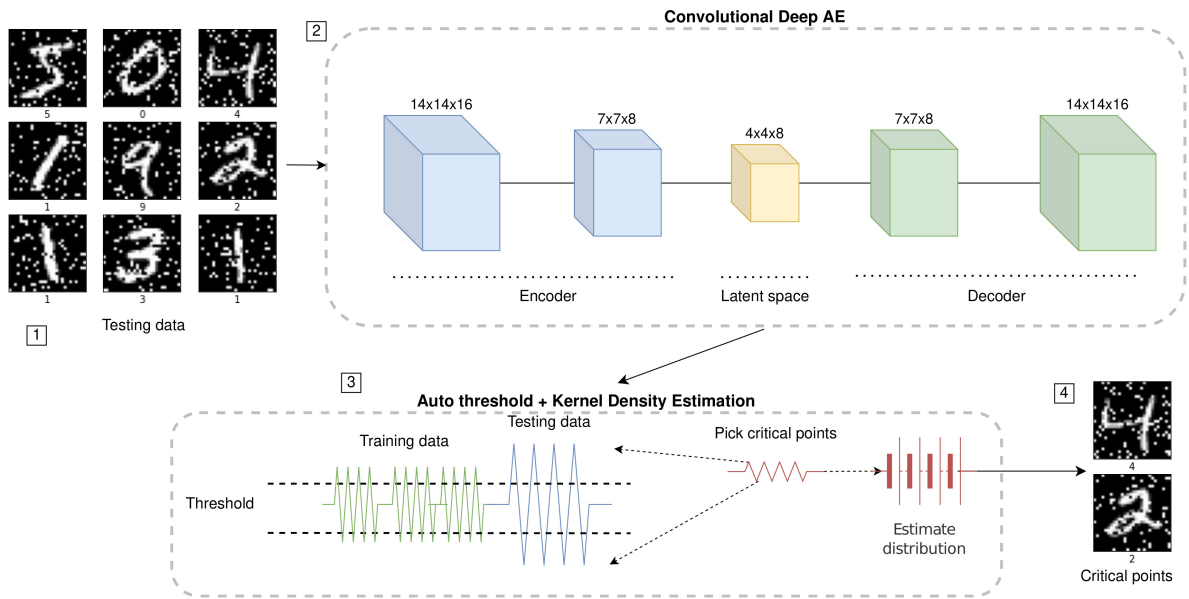

**Figure 14:** Architecture for concept drift detection and adaptation that comprises a Convolutional Deep AutoEncoder and a Critical Point Selector, which consists of an Auto-threshold and a Kernel Density Estimator. The Autoencoder contains 7 convolutional layers, 3 max-pooling layers, and 3 up-sampling layers. It learns from the original dataset and predicts on the new corrupted data. Then, the Auto-threshold calculates the best boundaries for test data. Finally, the Density Estimator picks the most likely critical points.

that are given to him or her, that is, the new critical values (X_test) and tag them. The objective is that the human agent uses the least effort possible since human time is a very scarce and expensive resource. For this purpose, this node is connected to the Drift Detector Checker to reduce the amount of observations to be tagged (only the most critical). This action improves notably the model score, as shown in the Experiments Section. The rationale behind this is that both learning algorithms and symbolic knowledge should work together to improve the model robustness. Algorithm 8 shows the procedure for this node, which follows the blueprint of a Checker node, as described before.

#### 3.4.4. Human Finder Checker
Similar to the Human Labeler Checker, it involves a human that works as a participatory agent. As shown in Algorithm 9, his/her job is to find the most similar observations on the training data given the critical values (X_test) from the Drift Detector Checker.

#### 3.4.5. Model Trainer Improver
An Improver node is in charge of applying feedback from previous nodes to enhance the current model. In this case, this node makes a simple retraining of the model using the training data plus the data tagged by the human agent in the previous step aiming to improve the model score. Following the blueprint of an Improver node, this procedure is shown in Algorithm 11.

#### 3.4.6. Model Denoiser Wrapper Improver
Having the pair of observations (X_clean, X_corrupted) returned by the Human Finder Checker, this Improver node trains a Convolutional Deep Denoising AutoEncoder to remove the noise from the corrupted data, as shown in Algorithm 10. Then, it wraps the current model (predictor) with the denoiser node, so that the result is a compound node (model + denoiser). This denoiser acts as a filter (to remove noise from the data) before making predictions to improve the robustness of the whole system. These steps are shown in Algorithm 12.

### 4. Experiments
In order to demonstrate the usefulness of Scanflow we provide experiments on how the interaction between special nodes (Tracker, Checker, and Improver) helps in the debugging of ML workflows through a human-in-the-loop approach.

#### 4.1. Testbed
For the experiments, we considered the following testbed:

- Platforms: 1) Ubuntu Linux 18.04 (64 bits); 2) Google Colab.

- Hardware: 1) Intel(R) Core(TM) i7-7600U CPU @ 2.80GHz, 16 GB RAM; 2) Intel(R) Xeon(R) CPU @ 2.30GHz, Nvidia K80 / T4, 12 GB RAM.

- Software: Anaconda3 Python 3.8[2] (Python data science distribution), Docker 19.03.12[3] (container manager), MLFlow 1.10 (streamline Machine Learning

---
[2]https://www.anaconda.com/
[3]https://www.docker.com

**Algorithm 7:** Critical Point Selector

**Input:** *autoencoder*, convolutional deep autoencoder;
**Input:** *X_train*: training data;
**Input:** *X_test*: testing data;
**Input:** *n_critical*: number of critical points;
**Input:** *h*, *K*: Bandwidth, Kernel;
crit_points ← { };
width ← 0.1;
quantile_high ← 0.80;
quantile_low ← 1.0 - quantile_high;
// Auto threshold
**for** *trials in range(3)* **do**
    loss_train ← get_loss(*autoencoder*, X_train)
    THRESHOLD_HIGH ← loss.quantile(q=quantile_high)
    THRESHOLD_LOW ← loss.quantile(q=quantile_low)
    loss_test ← get_loss(*autoencoder*, X_test)
    high ← loss_test > THRESHOLD_HIGH
    low ← loss_test < THRESHOLD_LOW
    bool_anomalies ← high | low
    **if** *length(X_critical) >= n_critical* **then**
      | break
    **end**
    quantile_high ← quantile_high - width
    quantile_low ← 1.0 - quantile_high
**end**
X_critical ← X_test[bool_anomalies]
// Kernel density estimator
$\widehat{f}_h(x) \leftarrow \frac{1}{nh} \sum_{x \in X} K\left(\frac{x-x_i}{h}\right)$
crit_points[$\widehat{x}$] ← sample($\widehat{f}(x)$, X_critical)
**if** *n_critical < length(crit_points)* **then**
    | return crit_points[:n_critical]
**end**
return crit_points
**Output:** crit_points: the most critical points;

**Algorithm 8:** Human Labeler Checker

**Input:** *X_test*: new critical observations;
feedback ← *None*;
// Prepare new data
X_test_prep ← preprocessing(X_test);
// Human intervention
labels ← human_labeler(X_test_prep);
// Save labels for retraining the model
feedback ← labels;
return feedback;
**Output:** *feedback*: information for improvement;

**Algorithm 9:** Human Finder Checker

**Input:** *X_train*: training data, clean values;
**Input:** *X_test*: most critical corrupted values;
feedback ← *None*;
// Prepare new data
X_corrupted ← preprocessing(X_test);
// Human intervention
X_clean, X_corrupted ← human_finder(X_train, X_corrupted);
// Save observations to create the denoisers
feedback ← X_clean, X_corrupted;
return feedback;
**Output:** *feedback*: information for improvement;

**Algorithm 10:** Convolutional Deep Denoising AutoEncoder

**Input:** *X*: clean data;
**Input:** *X̂*: corrupted data;
**Input:** *epochs*, *batches*, *lr*: Number of epochs, number of batches, learning rate;
**Input:** *W*, *b*, $b_h$: Parameters ($\theta$) : Weights, bias, latent bias;
**for** *epoch in epochs* **do**
    **for** *batch in batches* **do**
        $h \leftarrow \text{relu}(\hat{X} * W + b)$
        $r \leftarrow \text{sigmoid}\left(h * W^T + b_h\right)$
        $L(x, r) \leftarrow -\sum_{i=1}^{d}\left[x_i \log r_i + (1-x_i)\log(1-r_i)\right]$
        $g \leftarrow \text{gradient}(\theta)$
        **for** $\theta_i, g_i$ *in* $(\theta, g)$ **do**
            | $\theta_i \leftarrow \theta_i - l * g_i$
        **end**
    **end**
**end**
autoencoder ← join(h, r)
return autoencoder
**Output:** *autoencoder*: convolutional deep denoising autoencoder;

development), Tensorflow 2.0[4] (Machine Learning library), and Pytorch 1.6.0[5] (Machine Learning library).

- Datasets: MNIST ([dataset] Feiyang Chen, Chen, Mao and Hu, 2018) (60,000 28×28 pixel grayscale images of handwritten digits from 0 to 9), MNIST-C ([dataset] Norman Mu and Gilmer, 2019) (handwritten digit database with 15 corruptions: corrupted version of MNIST), CIFAR-10 ([dataset] Alex Krizhevsky, 2009) (60,000 32x32 colour images in 10 classes), CIFAR-10-C ([dataset] Dan Hendrycks and Dietterich, 2019) (corrupted version of CIFAR-10), FashionMNIST ([dataset] Han Xiao, Rasul and Vollgraf, 2017) (60,000 28x28 pixel grayscale images of 10 types of clothing), and FashionMNIST-C ([dataset] Dan Hendrycks and

---
[4]https://www.tensorflow.org
[5]https://www.pytorch.org

**Algorithm 11:** Model Trainer Improver

**Input**: $m$: model;
**Input**: $X\_feedback$, $y\_feedback$: new labeled critical observations;
**Input**: $X\_train$, $y\_train$: training data;
**Input**: $q$: accuracy metric;
```
// Retrain the current model with augmented data
```
$m' \leftarrow$ retrain($m$, X_train+X_feedback, y_train+y_feedback);
```
// Get score from modified model
```
$q' \leftarrow$ evaluate($m'$);
```
// If new model is better, save it, else, do nothing
```
**if** $q' > q$ **then**
  | return $m'$, $q'$;
**end**
return None, None;
**Output**: $m'$: new model;
**Output**: $q'$: new accuracy;

---

**Algorithm 12:** Model Denoiser Wrapper Improver

**Input**: $m$: model;
**Input**: $X\_feedback$, $y\_feedback$: new corrupted observations with its corresponding clean version from training;
**Input**: $X\_train$, $y\_train$: training data;
**Input**: $q$: accuracy metric;
X_clean, X_corrupted $\leftarrow$ X_feedback
```
// Retrain the current model with augmented data;
```
$m' \leftarrow$ retrain($m$, X_train+X_corrupted, y_train+y_feedback)
```
// Get the denoiser from the feedback;
```
denoiser $\leftarrow$ conv_denoise_autoencoder(X_clean, X_corrupted);
```
// Wrap (append) the modified model with a denoiser;
```
$m'' \leftarrow$ wrap($m'$, denoiser);
```
// Get score from modified model;
```
$q'' \leftarrow$ evaluate($m''$);
```
// If new model is better, save it, else, do nothing;
```
**if** $q'' > q$ **then**
  | return $m''$, $q''$;
**end**
return None, None;
**Output**: $m''$: new model;
**Output**: $q''$: new accuracy;

---

Dietterich, 2019) (corrupted version of FashionMNIST).

- Container's environment: Debian Linux 10.0[6], Miniconda 3[7].

---
[6]https://www.debian.org/releases/stable/
[7]https://docs.conda.io/en/latest/miniconda.html

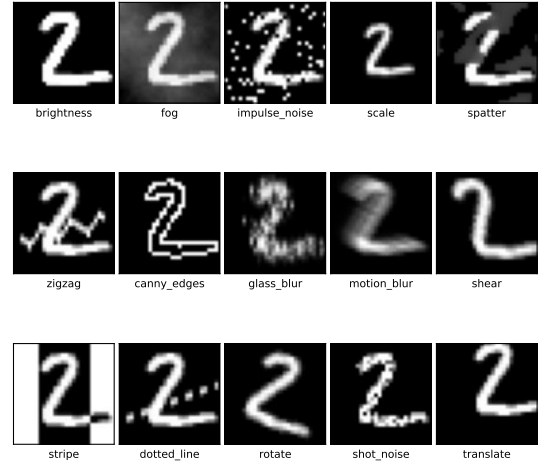

**Figure 15**: MNIST-C dataset. This figure shows the 15 corruptions this dataset applies to the original MNIST dataset.

### 4.2. Experiment 1: Human-in-the-loop ML model debugging on MNIST

This set of experiments evaluates the special nodes (Checker, Tracker, and Improver) in practice using a real use case to demonstrate human-in-the-loop and debugging features. We built a workflow that contains a digit classifier that uses the well-known hand-written digits dataset MNIST and, in particular, its derived version MNIST-C with 15 corruptions, which are shown in Figure 15. We chose this dataset due to its fast training, its high relevance for machine learning (Frosst, Sabour and Hinton, 2018; Schott, Rauber, Bethge and Brendel, 2018), and the option to compare against related works using it ([dataset] Norman Mu and Gilmer, 2019). The modeling architecture for this dataset was built as a Convolutional Neural Network (CNN) (Lecun and Bengio, 1995).

Five scenarios were tested to proof the usefulness of using learning and symbolic special nodes in the Scanflow graph structure. Each scenario is evaluated on 15 corruptions (MNIST-C) and with different human interventions, that is, labeling 10, 25, 50, 100, 200, and 500 observations. The following description explains these scenarios:

- Scenario 1 (Baseline): This scenario does not involve any modification, it just trains a model on the original MNIST, which is then evaluated on both the MNIST and the MNIST-C datasets. That is to say, Baseline = Train(X_train, y_train) and then Evaluate(X_test) and Evaluate(X_test_c). These two scores are the baseline marks for the following scenarios.

- Scenario 2 (Naive): This scenario takes the baseline model trained on the original MNIST to predict the

new corrupted points (from MNIST-C), and then uses these new samples to augment the data and retrain the model. That is, Retrain(X_train + X_new_c, y_train + y_pred_c). Figure 16 shows the pipeline for this scenario.

- Scenario 3 (DDC+HLC): DDC stands for Drift Detector Checker and HLC for Human Labeler Checker. This scenario uses the Drift Detector Checker to select the most critical points (X_critical), asks a human to label them (y_labels), and then uses these new samples to augment the data and retrain the model. That is, Retrain(X_train + X_critical, y_train + y_labels). This scenario combines symbolic (Human Labeler Checker, Critical Point Selector) and learning algorithms (Convolutional Deep Autoencoder). Figure 17 shows the pipeline for this scenario.

- Scenario 4 (RND+HLC): RND stands for Random. This scenario is similar to Scenario 3, but without using the Drift Detector Checker to select critical points. Instead, it uses a random choice to take only *n_critical* observations (X_random). Then, as in Scenario 3, these random observations are labeled by a human (y_labels_random) and used to retrain the model, so that Retrain(X_train + X_random, y_train + y_labels_random). The goal of this scenario is to show how good Scenario 3 is compared to a random selection of critical points. Figure 18 shows the pipeline for this scenario.

- Scenario 5 (MDWI+HFC): MDWI stands for Model Denoiser Wrapper Improver and HFC for Human Finder Checker. This scenario starts like Scenario 3 by using the Drift Detector Checker to select the most critical points (X_critical). However, the human agent has a different role as now he/she has to find on the training data the most similar observations to the most critical points received from the Drift Detector Checker. Using this information, a denoising autoencoder is created to be appended to the current model. This allows the model to filter out the noisy observations before making predictions. Figure 19 shows the pipeline for this scenario.

The methodology to compare these scenarios is as follows:

- First, test the new model on the MNIST dataset. This evaluation shows how the model performs on the original distribution.

- Second, test the new model on the MNIST-C dataset (average over the 15 corruptions). This evaluation represents the model performance on unseen data that has drifted from the original distribution.

- Average these two scores to get the overall mean accuracy. This evaluation depicts the model generalization on both the original and new data.

Figure 20 shows the accuracy of all the scenarios on the MNIST dataset. All the scenarios but the Naive one obtain nearly the same accuracy than the baseline for this dataset ($\pm$ 0.08). Here, the Naive approach performs slightly worse because some of the predictions that are used as labels are wrong (the model is not perfect), and this degenerates the baseline model when these samples are included in the training set. Note that DDC+HLC and MDWI+HFC have exactly the same performance because they are using the same model (there is not noise to filter in the MNIST dataset, and therefore the MDWI component has not effect).

Figure 21 shows the accuracy of all the scenarios on the MNIST-C dataset and Figure 22 shows the average accuracy on both datasets, which is pretty similar to the former since the small score variation in Figure 20 does not affect much to the resulting average. As shown in these figures, the evaluation on the MNIST-C dataset shows a completely different behavior. Here, scenarios DDC+HLC, RND+HLC, and MDWI+HFC perform significantly better than the baseline, with scenarios DDC+HLC and MDWI+HFC (the ones proposed in this paper) performing the best. As shown in the score table, the Naive scenario has considerably worse performance and for this reason it has been excluded from the plot to avoid problems with Y axis scaling. DDC+HLC clearly outperforms RND+HLC when the number of labeled observations is small. This confirms the advantage of using the DDC component to select the most critical observations that must be labeled to provide the highest benefit to the model. As the number of labels increases, the two approaches tend to converge, because when using a random selection approach, the amount of samples to be labeled increasingly coincides with the critical points, since there is a fixed amount of runtime observations for both scenarios. Therefore, the DDC approach will be very useful for cases where there is little availability of human agents.

MDWI+HFC shows the best performance, demonstrating the benefit that results when combining the special nodes (drift detector checker, model denoising autoencoder, and human interventions) with a traditional workflow. Although MDWI+HFC also tends to converge with the *-HLC approaches as the number of observations increases, this has a different cause. MDWI+HFC creates a denoiser for each type of corruption, which helps to better identify the noise and eventually remove it before doing the predictions, but these denoisers fail when the corruption is of the translational or scaling types. These corruptions do not add a layer of noise, but shift and rotate the images. Those changes are not learned correctly by the denoisers, and as the number of observations increases, they create increasingly erroneous representations for these two types of noise, decreasing the growth rate in the model performance.

### 4.3. Experiment 2: Human-in-the-loop ML model debugging on CIFAR-10 and FashionMNIST

This experiment demonstrates the ability of Scanflow to support more diverse datasets and more complex modeling architectures. For that purpose, we developed models for two additional datasets. The first dataset is CIFAR-10 (and its corresponding corrupted version), which contains colour images of motor vehicles and animals. Figure 23a and Figure 23b show samples of CIFAR-10 and CIFAR-10-C, respectively. The modeling architecture for this dataset was built on top of DenseNet (Huang, Liu, van der Maaten and Weinberger, 2018) and using transfer learning on ImageNet weights (Russakovsky et al., 2015). The second dataset is FashionMNIST (and also its corresponding corrupted version), which contains grayscale images of clothing. Figure 27a and Figure 27b show samples of FashionMNIST and FashionMNIST-C, respectively. Differently from the MNIST dataset, we trained FashionMNIST dataset on a fully connected Multi Layer Perceptron (Haykin, 1994).

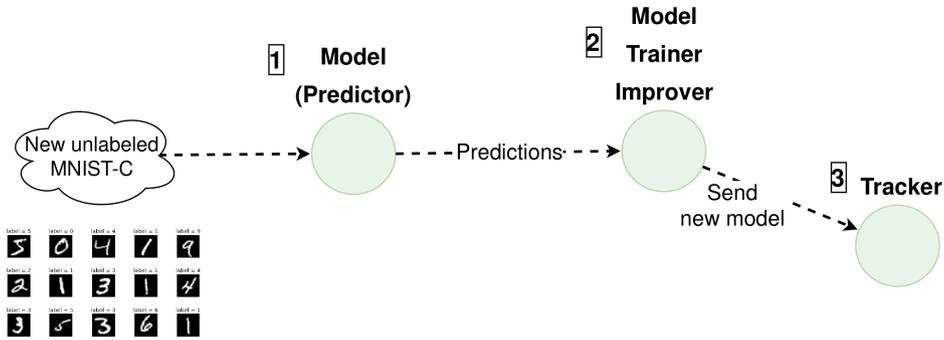

**Figure 16:** Pipeline for scenario 2. It comprises a Model (current predictor), a Model Trainer Improver (only to retrain the model) and a Tracker (to save metadata, in this case, the new model).

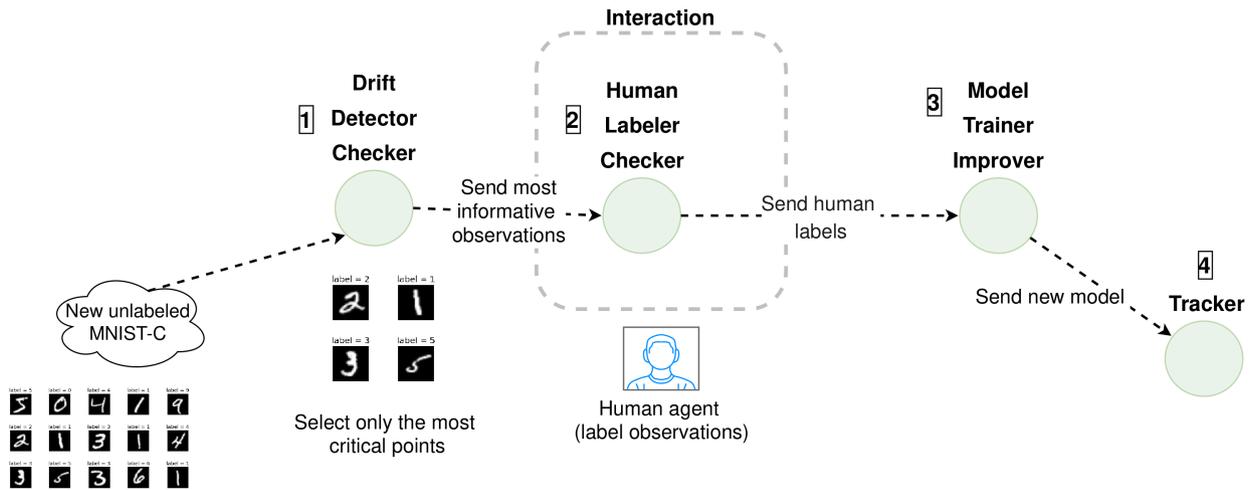

**Figure 17:** Pipeline for scenario 3. It involves a Drift Detector Checker (to detect anomalous observations), a Human Labeler Checker (to label those observations), a Model Trainer Improver (to retrain the current model), and a Tracker (to save the new model).

We tested scenarios Baseline, Naive, DDC+HLC, and RND+HLC, as described in the previous experiment, and we used the same methodology to compare the results. As shown in the score tables, the Naive scenario has always considerably worse performance and for this reason it has been excluded from the plots to avoid problems with Y axis scaling.

Figure 24 and Figure 28 shows the accuracy of the tested scenarios on the original CIFAR-10 and FashionMNIST datasets, respectively. All the scenarios obtain nearly the same accuracy than the baseline for this dataset (± 0.6), although the accuracy of DDC+HLC and RND+HLC on CIFAR-10 degrades minimally with more labels, because retraining the model with the new labels adapts it to predict better the corrupted dataset (so the accuracy improvement in this case is higher, as shown below).

Figure 25 and Figure 29 show the accuracy of the tested scenarios on the CIFAR-10-C and FashionMNIST-C corrupted datasets, respectively, and Figure 26 and Figure 30 show the average accuracy on both datasets (original+corrupted), which is pretty similar to the corrupted one. The plots show that scenarios DDC+HLC and RND+HLC clearly outperform the Baseline accuracy, that using more labels tends to improve the model accuracy, and that the model generalizes better when using DDC+HLC rather than RND+HLC, because the drift detector yields better significant points to be labeled.

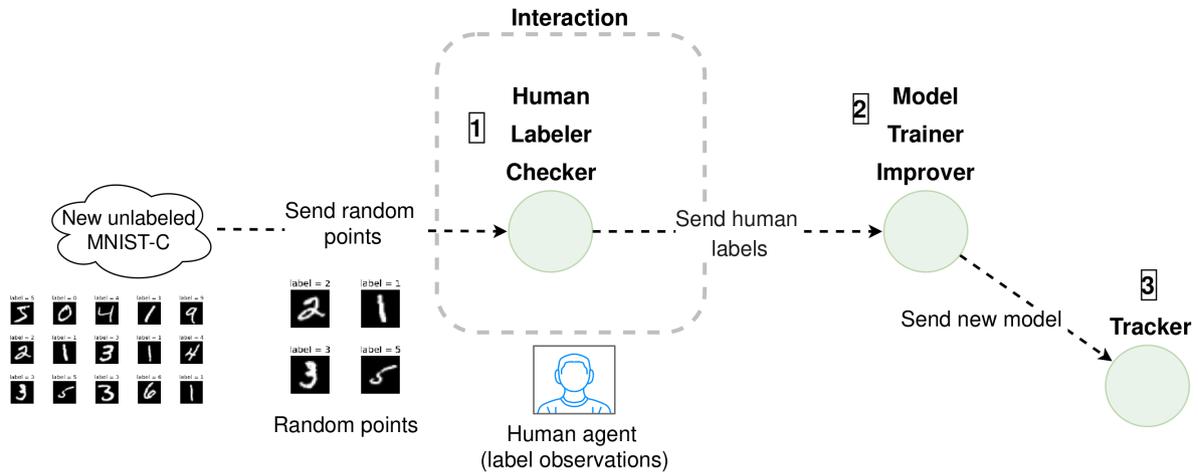

**Figure 18:** Pipeline for scenario 4. It includes a Human Labeler Checker, a Model Trainer Improver and a Tracker. Similar to scenario 3, but without a Drift Detector Checker.

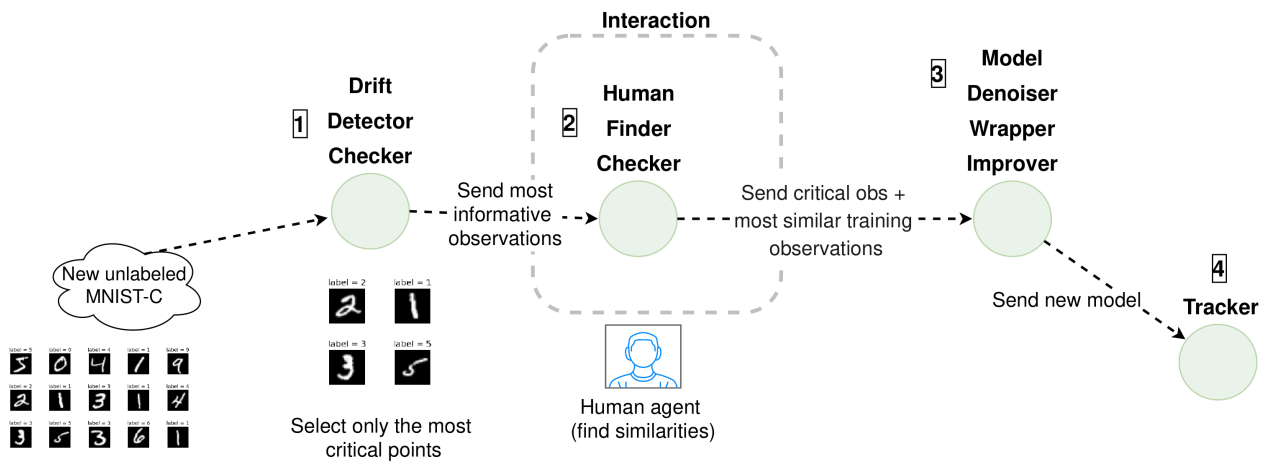

**Figure 19:** Pipeline for scenario 5. It contains a Drift Detector Checker (to detect anomalous observations), a Human Finder Checker (find similarities of the anomalous observations on the training data), a Model Denoiser Wrapper Improver (which uses the similarities to create a denoising autoencoder), and a Tracker (to save the new model: denoiser + current model).

## 5. Related work

In recent years, several technologies have emerged to facilitate the work with workflows in general, such as Airflow[8] and Luigi[9], as well as those applied specifically to Machine Learning, such as Kubeflow[10], MLflow, and Seldon[11]. It is evident that there is a need for tools to create simple, portable, and scalable pipelines that allow easy ML deployments for various infrastructures. For instance, Kubeflow, Airflow, MLflow, and Luigi deal with deployment, dependency resolution, and tracking and monitoring of workflows, with more or less intensity. Similarly, Seldon includes those features and also deals with explainability and outlier/drift detection[12]. However, current ML models need not only a passive behavior, that is to say, to have a good deployment, to get nice explanations, and to detect critical data instances, but also to be actively reformulated and updated in an always changing world.

Perhaps, the key difference between these frameworks and Scanflow is that the latter, in addition to debugging, also performs model updating with human intervention, which of course, not only makes the model better but also the original data (by providing good labels).

On the other hand, different methods to deal with the lack of robustness in ML have been proposed. In addition to the concept drift problem addressed in this paper (see the corresponding related work in Section 3.4.1), other problems that compromise model generalization include for instance adversarial attacks. For example, Xie, Wu, van der

---

[8] https://airflow.apache.org/
[9] https://github.com/spotify/luigi
[10] https://www.kubeflow.org/
[11] https://www.seldon.io/
[12] https://www.seldon.io/tech/products/alibi/

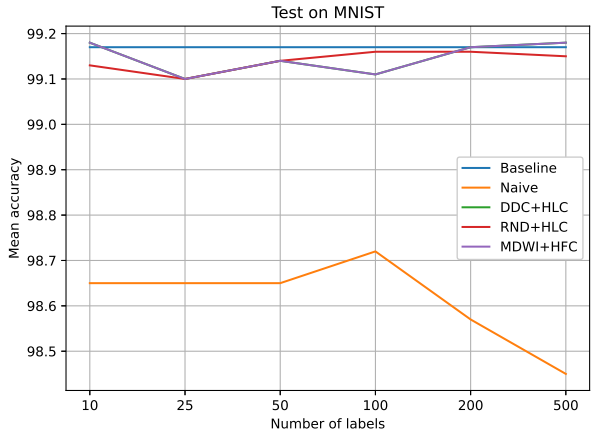

(a) Mean accuracy plot

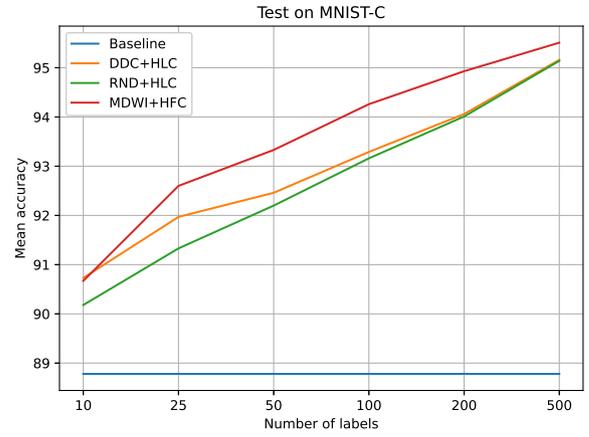

(a) Mean accuracy plot

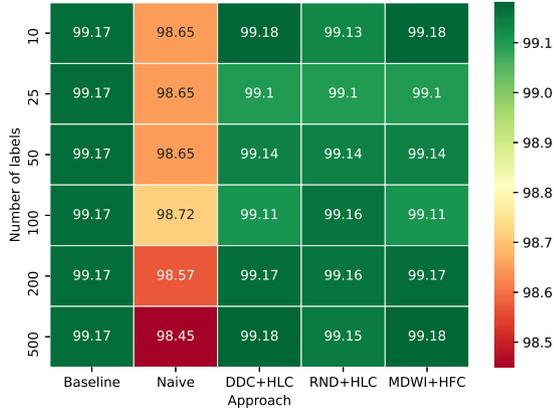

(b) Mean accuracy score table

**Figure 20:** Evaluation on MNIST original dataset for each scenario.

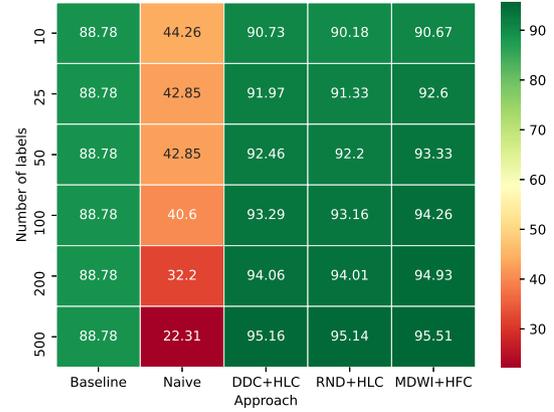

(b) Mean accuracy score table

**Figure 21:** Evaluation on MNIST-C dataset (with the 15 corruptions) for each scenario.

Maaten, Yuille and He (2018) developed a feature denoising method to improve adversarial robustness for ImageNet (Deng, Dong, Socher, Li, Kai Li and Li Fei-Fei, 2009). Similarly, Huang and Zhang (2019) presented a method based on low-dimensional embedding using a pretrained model to alleviate this problem (that includes also not-learning steps). These works are related to our proposal as they also aim to improve robustness in models. However, our work does it externally without modifying the current code (e.g., without modifying the learning model). Our claim is our particular architecture to deal with robustness, that is to say, the external joint work between the Tracker, Checker, and Improver (special nodes) that can include other approaches (even better anomaly detectors) as nodes to improve models over time. Focusing on the MNIST dataset, many defensive methods have been applied recently on this regard, in particular by using the MNIST-C dataset to evaluate their performance ([dataset] Norman Mu and Gilmer, 2019). Table 1 compares those previous results with the approaches proposed in this paper. Whereas it is true that some of these methods can improve our accuracy in certain types of corruption, on average our methods provide better performance and therefore they generalize better.

**Table 1**
Mean accuracy of various models on MNIST-C

| Model | Mean Accuracy |
|---|---|
| Baseline | 88.78 |
| Conv2 (PGD) | 80.06 |
| Conv3 (PGD) | 78.86 |
| Conv3 (GAN) | 81.14 |
| Capsule | 77.60 |
| ABS | 82.46 |
| RND+HLC* | 91.33 |
| **DDC+HLC*** | **91.97** |
| **MDWI+HFC*** | **92.60** |

\* Involves human intervention. Best results are bolded

Finally, few research works have considered hybrid systems involving symbolic knowledge that aims at increasing

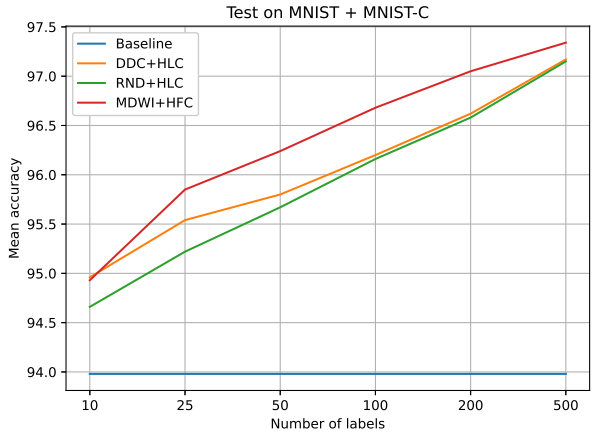

(a) Mean accuracy plot

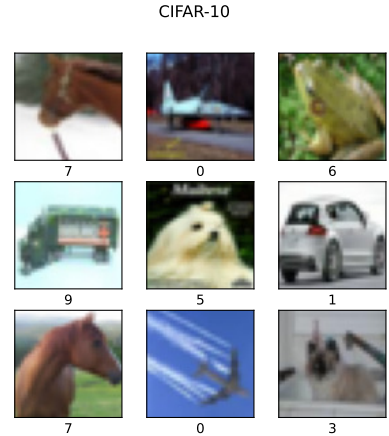

(a) Original CIFAR-10 test sample.

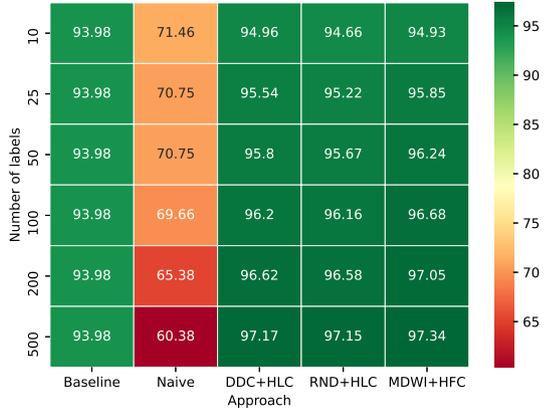

(b) Mean accuracy score table

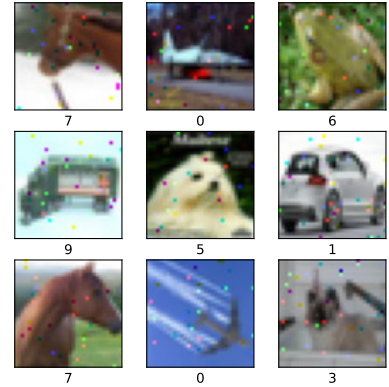

(b) CIFAR-10-C test sample with impulse noise corruption (severity 1).

**Figure 22:** Evaluation on MNIST + MNIST-C datasets for each scenario.

**Figure 23:** CIFAR-10 and CIFAR-10-C datasets. This figure shows 9 images and their respective corrupted version.

model generalization. An example is the Neuro-Symbolic Concept Learner (Mao, Gan, Kohli, Tenenbaum and Wu, 2019) developed by IBM and MIT, a system that combines the reasoning power of rule-based systems and the learning flexibility of neural networks to solve visual question answering (VQA) problems, which are difficult to be solved with only a learning approach. As a result, this system gets better results with less training data and yields explainable results. This study conducted a similar idea to our work in the way of combining approaches to come up with a hybrid architecture. Nonetheless, we consider wrappers (special nodes) which interact, externally, with the current workflow without altering, internally, its parts (nodes). Besides, Marcus (2020) presented a proposal of some steps to tackle this issue, such as creating hybrid neuro-symbolic architectures, developing cognitive models, and emerging reasoning mechanisms to operate on these models. Likewise, Chollet (2019) presented similar ideas on how to help the reasoning of tasks by means of certain strategies, on which combined with a learning method, lead to further steps towards robust artificial intelligence.

## 6. Conclusion and future work

In this paper, we have introduced Scanflow, a multi-graph containerized framework for ML workflow management, supervision, and debugging that seeks to provide initial steps towards improving the robustness of workflows in run time. For this purpose, it builds on a graph structure that offers high flexibility when operating on pipelines, as well as, on the work of special nodes that are responsible for detecting and ensuring the proper functioning of those workflows by using both learned and human-provided knowledge.

As for the structure, it enables flexible management allowing users to have a complete control of what is happening on each node, simplifying the deployment and helping the debugging task. Since Scanflow is built upon graph theory, it inherits graph properties that allow, for instance, finding node neighbors, successors, and predecessors, which are really useful to find dependencies (backbone ideas on causality), and also using a rich myriad of algorithms, such as, path searching, community detection, connected components, page-rank, etc. Apart of that, this architecture allows nodes to be lazily executed, that is, being triggered when an action is performed (e.g., when labeling, a rule-based logic, or any intervention).

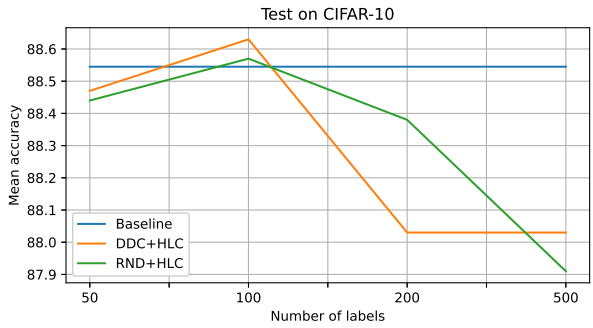

(a) Mean accuracy plot

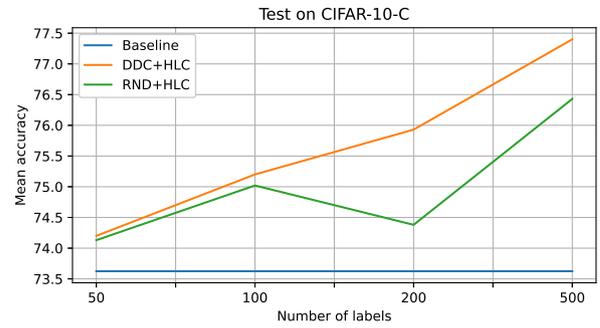

(a) Mean accuracy plot

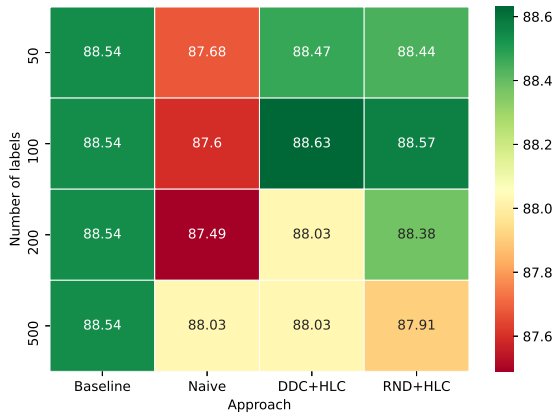

(b) Mean accuracy score table

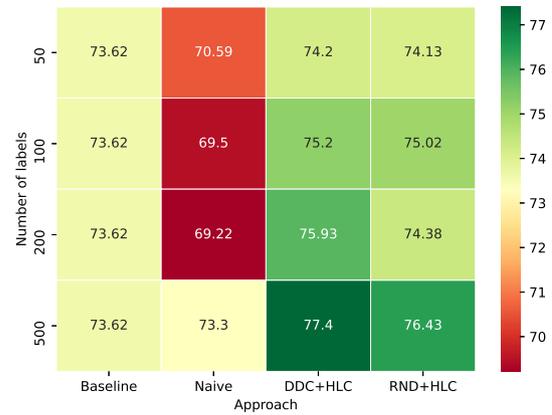

(b) Mean accuracy score table

**Figure 24:** Evaluation on CIFAR-10 original dataset for each scenario.

**Figure 25:** Evaluation on CIFAR-10-C dataset for each scenario.

As for the usage of special nodes for debugging, we have presented novel ideas on how to perform continuous improvements on ML models over time through the joint work of Tracker, Checker, and Improver nodes to track models metadata, check their behavior in production, and refine them. Additionally, we have provided built-in implementations of those nodes for detecting drift distribution, for selecting the most critical observations, for removing noise from data, and for including human agents in the debugging loop. That is to say, mechanisms to deal with not only ordinary or normal data distribution (Independent and Identically Distributed) but also with atypical or more general data distribution (Out-of-Order Distributed). Our experiments with MNIST-C, CIFAR-10, and FashionMNIST datasets have shown accuracy and robustness results that outperform other models with the help of human intervention.

As future work, we expect to work on improving the interface between symbolic and learning operations to accelerate debugging, focusing especially on the refinement of the human-computer interaction to get good heuristics and labels (domain knowledge) which can be leveraged to introduce ideas of data-centric techniques like weak supervision[13]. Besides, we can exploit further the compositionality feature of the framework, that implies that a user might know any node behavior over time, and as a result, a workflow could be explained as a whole by its parts. Naturally, the workflow might be explained by measuring the effects that data causes on it, externally. On the other hand, we will continue working on the creation of built-in nodes that can be generalized to many tasks and provide a common template for the resolution of many problems. Finally, we plan to integrate Scanflow with well-known container orchestrators, such as Kubernetes, to offer more deployment options to data scientists.

---

[13] http://ai.stanford.edu/blog/weak-supervision/

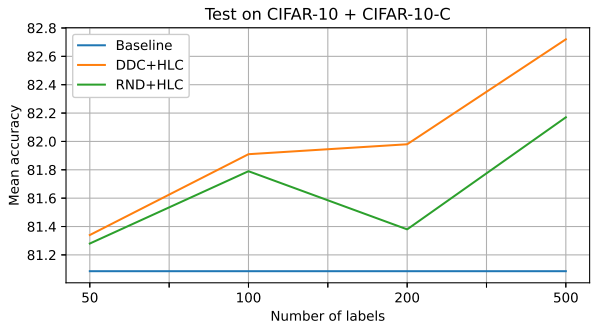

(a) Mean accuracy plot

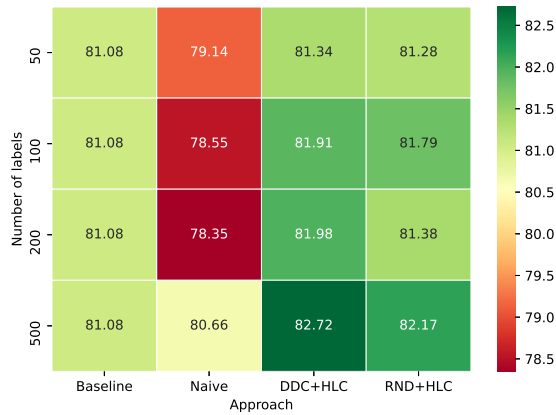

(b) Mean accuracy score table

**Figure 26:** Evaluation on CIFAR-10 + CIFAR-10-C datasets for each scenario.

## Declarations

### Funding

This work was partially supported by Lenovo as part of Framework Contract Lenovo-BSC 2020, by the Spanish Government under contract PID2019-107255GB, and by the Generalitat de Catalunya under contract 2017-SGR-1414 and under grant 2020 FI-B 00257.

### Competing interests

The authors declare that they have no competing interests.

### Availability of data and material

The datasets used to perform the current study are MNIST[14], MNIST-C[15], CIFAR-10[16], CIFAR-10-C[17], FashionMNIST[18], and FashionMNIST-C[19] (noise functions were applied on FashionMNIST to convert it to FashionMNIST-C because such dataset does not exist).

---

[14]http://yann.lecun.com/exdb/mnist/
[15]https://zenodo.org/record/3239543/
[16]https://www.cs.toronto.edu/ kriz/cifar.html
[17]https://zenodo.org/record/2535967#.YWQZZYBBxKQ
[18]https://github.com/zalandoresearch/fashion-mnist
[19]https://github.com/hendrycks/robustness

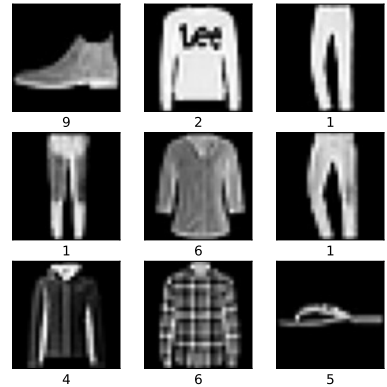

(a) Original FashionMNIST test sample.

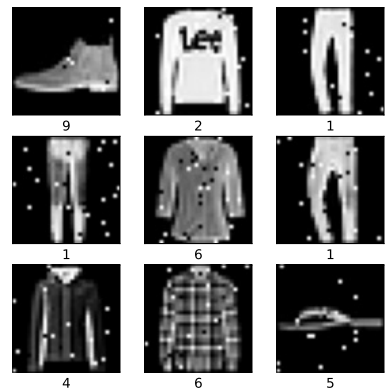

(b) FashionMNIST-C test sample with impulse noise corruption (severity 1).

**Figure 27:** FashionMNIST and FashionMNIST-C datasets. This figure shows 9 images and their respective corrupted version.

### Code availability

The code for reproducing the experiments is available in three Google Colab notebooks (one for each dataset), namely, experiment 1 (MNIST)[20], experiment 2 (CIFAR-10)[21] and experiment 2 (FashionMNIST)[22]. Moreover, the Scanflow implementation is available in GitHub[23].

### Authors' contributions

The first draft of the manuscript was written by Gusseppe Bravo-Rocca and Jordi Guitart and all authors commented on previous versions of the manuscript. All authors read and approved the final manuscript.

---

[20]shorturl.at/iluM2
[21]shorturl.at/dpuHQ
[22]shorturl.at/qtNOY
[23]https://github.com/gusseppe/scanflow

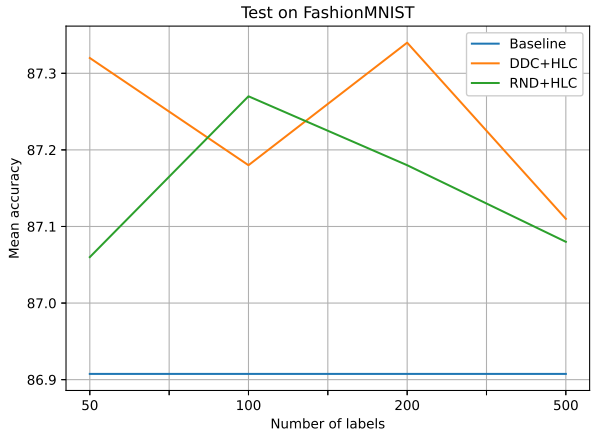

(a) Mean accuracy plot

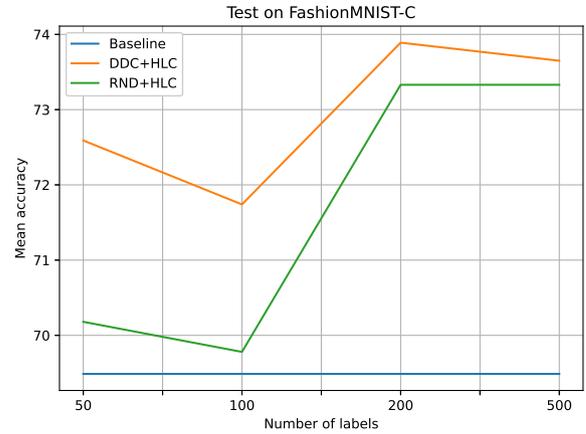

(a) Mean accuracy plot

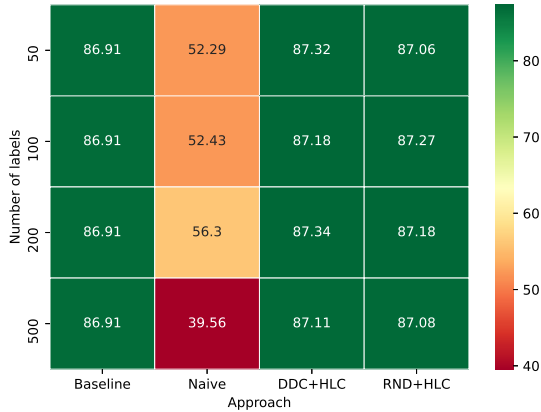

(b) Mean accuracy score table

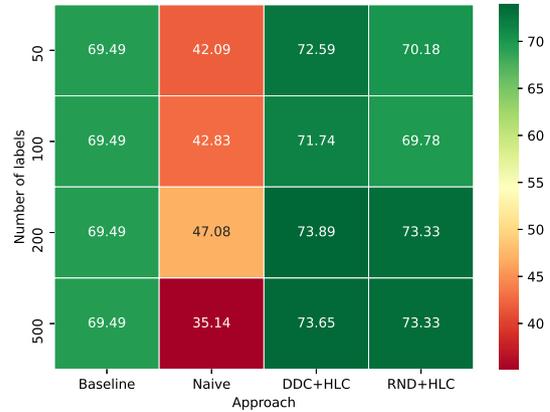

(b) Mean accuracy score table

**Figure 28:** Evaluation on FashionMNIST original dataset for each scenario.

**Figure 29:** Evaluation on FashionMNIST-C dataset for each scenario.

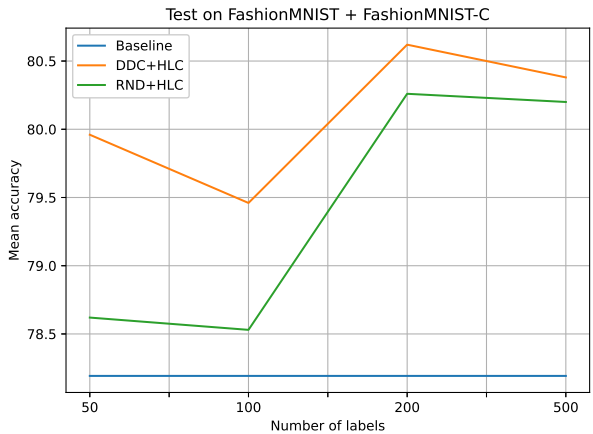

(a) Mean accuracy plot

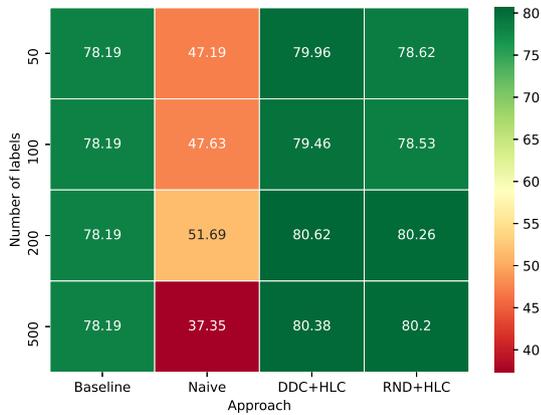

(b) Mean accuracy score table

**Figure 30:** Evaluation on FashionMNIST + FashionMNIST-C datasets for each scenario.